\documentclass{ngsm2024} %
\usepackage{mathtools}
\usepackage{booktabs}
\usepackage{bm}
\usepackage{algorithm}
\usepackage{algpseudocode}

\title{When can transformers compositionally generalize in-context?}

\optauthor{%
\Name{Seijin Kobayashi$^*$} \Email{seijink@ethz.ch}\\
\addr ETH Zürich
\AND
\Name{Simon Schug$^*$} \Email{sschug@ethz.ch}\\
\addr ETH Zürich
\AND
\Name{Yassir Akram$^*$} \Email{yakram@ethz.ch}\\
\addr ETH Zürich
\AND
\Name{Florian Redhardt} \Email{fredhardt@student.ethz.ch}\\
\addr ETH Zürich
\AND
\Name{Johannes {von Oswald}} \Email{jvoswald@google.com}\\
\addr Google, Paradigms of Intelligence Team
\AND
\Name{Razvan Pascanu} \Email{razp@google.com}\\
\addr Google DeepMind
\AND
\Name{Guillaume Lajoie} \Email{g.lajoie@umontreal.ca}\\
\addr Université de Montréal,
\addr Mila
\AND
\Name{João Sacramento} \Email{joaosacramento@google.com}\\
\addr Google, Paradigms of Intelligence Team}

\begin{document}

\maketitle
\vspace{-1.5cm}
\def\footnoteseptext{ }
\def\thefootnote{$^*$}\footnotetext{Equal contribution, alphabetical order.}\def\thefootnote{\arabic{footnote}}
\def\footnoteseptext{. }
\begin{abstract}%

Many tasks can be composed from a few independent components.
This gives rise to a combinatorial explosion of possible tasks, only some of which might be encountered during training.
Under what circumstances can transformers compositionally generalize from a subset of tasks to all possible combinations of tasks that share similar components?
Here we study a modular multitask setting that allows us to precisely control compositional structure in the data generation process.
We present evidence that transformers learning in-context struggle to generalize compositionally on this task despite being in principle expressive enough to do so.
Compositional generalization becomes possible only when introducing a bottleneck that enforces an explicit separation between task inference and task execution.
\end{abstract}


\section{Introduction}

Many tasks are compositional and as a result, there is a combinatorial explosion of possible tasks.
In this setting, when exposed to a number of tasks sharing components, it is desirable for a learning system to master operations that can be reused and leveraged to generalize to entirely new tasks.
Ideally, our learning systems could discover the constituent parts underlying the compositional task structure, and naturally generalize compositionally.
Prior work has explored this question~\citep{mittal2022is}, and recent results show that gradient-based meta-learning with hypernetworks can compositionally generalize after training only on a linear subset of tasks~\citep{schug_discovering_2024}.
Can transformers learning in-context achieve the same thing?
In-context learning is powerful, with evidence pointing towards it being able to implement mesa-optimization \citep{dai_why_2023, akyurek_what_2023, von_oswald_uncovering_2023}.
In some settings, compositional generalization appears to be in reach \citep{an_how_2023, lake_human-like_2023, hosseini_compositional_2022,schug2024attention}.
However, there are also instances where despite being able to identify latent task information, generalization fails \citep{mittal2024does}.
To shed light on the question under what circumstances transformers can learn to compositionally generalize in-context, we study the synthetic multitask setting previously introduced by \citep{schug_discovering_2024}.
We find that while the transformer is able to correctly infer the task latent variable, and solve in-distribution tasks, it fails to appropriately generalize.
Introducing a bottleneck into the architecture separating task inference and task execution helps to overcome this failure by enabling compositional generalization.
We design several interventions and decoding analyses that suggest that the success of the bottleneck is due to encouraging the discovery of the modular structure of the underlying task generative model.
This finding paves the way to possible architectural inductive biases that may promote better generalization in-context for transformers.

\section{Generating modular tasks with compositional structure}
\begin{figure}[h]
    \centering
    \includegraphics[width=\textwidth]{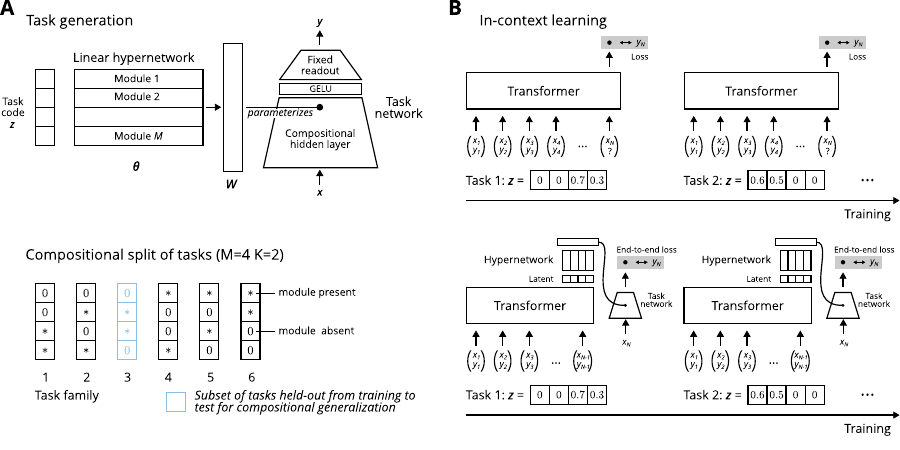}
    \caption{\textbf{In-context learning of compositional tasks.} \textbf{A} We create a modular task distribution with compositional structure using a linear hypernetwork which parameterizes a nonlinear task network (upper). To test for compositional generalization, we hold-out a subset of tasks of certain module combinations from training (lower). \textbf{C} We present each task, specified by a task latent code $\bm{z}$, in-context to a transformer that directly predicts the target of the query (upper) and to a transformer with a latent bottleneck that is used as the input to a learned hypernetwork (lower).}
    \label{fig:conceptual}
\end{figure}

\paragraph{Task generation} 
We aim to study a multitask setting where tasks share compositional structure.
To this end, we consider an adapted version of the synthetic setting introduced by \citep{schug_discovering_2024} which provides full control of the underlying compositional structure and allows us to measure the ability for compositional generalization.
Specifically, we will leverage a task-shared linear hypernetwork \citep{ha_hypernetworks_2017} that defines a regression task given a low-dimensional task code $\bm{z}$ as shown in Figure~\ref{fig:conceptual}A.
The linear hypernetwork is parameterized by a set of modules $\{\bm{\theta}_m\}_{m=1}^M$. Given $\bm{z}\in \mathbb{R}^M$, it produces task-specific parameters $\bm{W} = \sum_{m=1}^M z_m \bm{\theta}_m$, which are used to parameterize a one hidden layer MLP with GELU nonlinearity and fixed readout weights with a single output unit, $g:(\bm{x}, \bm{W}) \mapsto g(\bm{x}, \bm{W})$.
This task network is used to define a regression task, producing labels $y = g(\bm{x}, \bm{W})$ given randomly drawn inputs $\bm{x} \sim   \mathcal{U}(-\sqrt{3}, \sqrt{3})$.
As a result, each task is obtained through the additive composition of a set of $M$ modules, where each module corresponds to a full set of task network parameters.
We can use this structure to explicitly test for compositional generalization by holding out a subset of module combinations during training (see bottom of Figure~\ref{fig:conceptual}A).
At evaluation, performance measured on tasks seen during training is referred to as in-distribution, while performance on the held-out tasks is considered out-of-distribution (OOD).
See Appendix~\ref{app:task_generation} for more details on the task generation and Appendix~\ref{app:metric} on how we define OOD tasks.

\paragraph{In-context learning}

We present the tasks in-context to transformer-based sequence models as shown in Figure~\ref{fig:conceptual}B.
For each task, we sample a set of pairs $\{(\bm{x}_i, y_i)\}_{i=1}^N$, concatenate each pair as a vector and present them as tokens to the transformer models.
For the final query token, we mask the label $y_N$ and train the model to predict it using a standard mean-squared error loss.
We compare two models with each other.
The first is a standard decoder-only transformer trained to directly predict $y_N$ (for details please consider Appendix~\ref{app:architecture}). 
We compare it to a transformer whose outputs are fed to a linear hypernetwork which parameterizes a single hidden layer task network that predicts the target of the query token, mirroring the generative process of the task.
This is an instance of a hypertransformer \cite{pmlr-v162-zhmoginov22a}.
While this model is still trained end-to-end using a mean-squared error loss on the target, the transformer can now specialize to infer a latent code and the parameters of the hypernetwork can be used to learn to execute the task given the latent code. 

\section{In-context compositional generalization}

As a consequence of the compositional structure, the number of possible tasks grows exponentially with the number of modules $M$.
Naively learning every combination independently therefore quickly becomes unfeasible.
Instead, ideally, our learning systems discover the underlying modular structure of the task while being exposed only to demonstrations of each task - not observing the ground-truth latent structure.
In the following experiments, we evaluate out-of-distribution (OOD) performance on held-out module combinations on the modular task described above consisting of $M=6$ modules.

\begin{figure}[h]
    \centering
    \includegraphics[width=\textwidth]{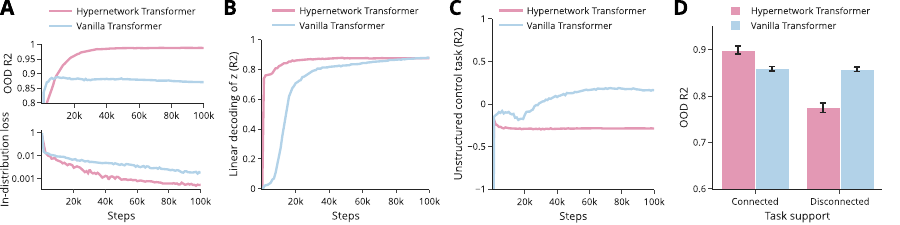}
    \caption{\textbf{In-context compositional generalization.} \textbf{A} The vanilla transformer fails to generalize compositionally to OOD tasks whereas the hypernetwork transformer generalizes (top). The vanilla transformer has a larger loss in-distribution compared to the hypernetwork transformer (bottom). \textbf{B} In both models it is possible to linearly decode the ground truth latent code from the residual stream of the transformer. \textbf{C} Testing both models on an unstructured control task unrelated to the compositional structure of the training distribution reveals that the hypernetwork transformer fails to perform unrelated tasks whereas the vanilla transformer retains some of these abilities.
    \textbf{D} \citep{schug_discovering_2024} suggests that when appropriately capturing the task generative model, compositional generalization is only possible given connected task support during training. While the OOD performance of a vanilla transformer is unaffected by this intervention, the ability of the hypernetwork transformer to compositionally generalize severely degrades for disconnected task support. }
    \label{fig:results}
\end{figure}

\paragraph{Transformers learning in-context fail to generalize compositionally.}

We first consider the vanilla transformer trained to predict the query label after observing examples of each task in-context.
As can be seen in Figure~\ref{fig:results}A, while being able to fit the in-distribution data relatively well (also compared to Figure~\ref{appfig:test-r2}), it fails to compositionally generalize to the held-out OOD tasks.
Surprisingly, despite this, Figure~\ref{fig:results}B  shows that it is possible to linearly decode the task latent code on OOD tasks from the residual stream given a linear decoder that is solely trained on in-distribution tasks (c.f. Appendix~\ref{app:metric} for more details).
This suggests that the model is able to implicitly perform task inference in a way that generalizes compositionally, yet it is unable to leverage the inferred $\bm{z}$ to predict the correct label on OOD tasks.

\paragraph{Separating task inference from task execution enables compositional generalization.}

Motivated by this observation, we equip the transformer with an explicit, learnable hypernetwork that takes as input the logits of the transformer as described above.
This encodes a strong architectural prior to separate task inference and task execution.
Indeed, Figure~\ref{fig:results}A shows that this system is able to compositionally generalize to held-out tasks while maintaining the linear decodability of the task latent code from the residual stream of the transformer (Figure~\ref{fig:results}B).

\paragraph{The vanilla transformer fails to capture the compositional structure of the task.}

To illuminate to what extent the two models discover the compositional structure underlying the task generative model, we perform two additional experiments.
First, we present both models with a control task that is also generated by a single hidden layer task network as used to produce the training tasks but crucially the parameters of this task network are not composed of the training modules but randomly initialized (see Appendix~\ref{app:random-control-task} for details).
Figure~\ref{fig:results}C shows that the hypernetwork transformer completely fails to solve this task over the course of training while the vanilla transformer displays modest performance, providing evidence that the former is strongly specialized to the particular compositional structure of the training tasks.
To complement this analysis, we construct two task distributions (connected vs disconnected) for training that have been shown by \citep{schug_discovering_2024} to causally affect the ability of hypernetworks to compositionally generalize (see Appendix~\ref{app:connectedness} for details).
Indeed the hypernetwork transformer is highly sensitive to this intervention while the vanilla transformer is virtually unaffected.
Taken together this suggests that the vanilla transformer fails to emulate the compositional structure of the task generative model which the hypernetwork transformer can capture.
This is striking noting that the former is in principle sufficiently expressive to implement the hypernetwork solution, see Appendix~\ref{app:hnet_construction} for an explicit construction.

\section{Discussion}
Despite the success of transformers at scale trained on simple autoregressive next-token prediction, there are still many failure cases - compositional generalization being one of them.
While we find that transformers are expressive enough to in principle encode compositional structure in a multitask setting, less powerful shortcuts dominate the solutions practically found by gradient-based optimization.
Encoding inductive biases into the architecture might help overcome these problems but finding such biases that generally work well is an open challenge.
Our results show that architecturally separating task inference from task execution through a bottleneck improves compositional generalization in our synthetic setting.
Future work should explore to what extent similar architectural motifs allow end-to-end discovery of compositional structure from data for a general class of problems.

\bibliography{bibliography}
\newpage
\clearpage
\appendix

\section{Implementing a hypernetwork in a transformer}\label{app:hnet_construction}

\subsection{Linear attention block}

We first give a brief overview of the linear transformer architecture.

Given input tokens $\bm{E} \in \mathbb{R}^{L \times d_m}$ for a sequence of length $L$, a transformer block consists of a self-attention layer followed by a multi-layer perceptron (MLP).
The transformation is done by first computing queries, keys and values $\bm{Q}, \bm{K}, \bm{V} = \bm{EW}_q, \bm{EW}_k, \bm{EW}_v $ with which we then update $\bm{E}$ as
\begin{align}
 \bm{E} & \gets \bm{E} + \bm{QK}^T\bm{V W}_P \\ 
 \bm{E} & \gets \bm{E} + \sigma(\bm{EW}_1)\bm{W}_2
\end{align}

where $\bm{W}_q, \bm{W}_k, \bm{W}_v \in \mathbb{R}^{d_m \times d_k}$ and $\bm{W}_p \in \mathbb{R}^{d_k \times d_m}$ as well as $\bm{W}_1 \in \mathbb{R}^{d_m \times d_h}, \bm{W}_2 \in \mathbb{R}^{d_h \times d_m}$ are learnable parameter matrices and $\sigma$ is a nonlinearity applied row-wise.
In practice, there are $H$ heads that perform the first attention operation in parallel, each with its own parameters $\bm{W}_q^{(h)}, \bm{W}_k^{(h)}, \bm{W}_v^{(h)}, \bm{W}_p^{(h)}$ for all $h$, resulting in the following forward function
\begin{align}
 \bm{E} & \gets \bm{E} + \sum_h^H \bm{Q}^{(h)}\bm{K}^{(h)\top}\bm{V}^{(h)} \bm{W}_P^{(h)} \\ 
\end{align}

\subsection{Construction}

We will now provide a construction of how a linear transformer can implement a fixed hypernetwork in the forward pass given any input $\bm{x} \in \mathbb{R}^d$ and latent $\bm{z}\in\mathbb{R}^M$.

\paragraph{Hypernetwork}
Let us consider the following linear hypernetwork:
\begin{equation}
    \bm{x},\bm{z} \rightarrow \bm{A} \sigma(\bm{W}(\bm{z}) \bm{x})
\end{equation}
 where \(\bm{W}(\bm{z}) = \sum_{m=1}^M z_m\bm{\theta}_m\), $\bm{\theta}_m \in \mathbb{R}^{h \times d} $ for all $m$ and $\bm{A} \in \mathbb{R}^{o \times d}$.

\paragraph{Token construction} We assume there are only $2$ tokens, $\bm{e}_1 = (\bm{x}^\top,\bm{0}_M,\bm{1}_{h+o})^\top$ and $\bm{e}_2 = (\bm{0}_d,\bm{z}^\top,\bm{0}_{h+o})^\top$ where $\bm{0}_k, \bm{1}_k$ indicate the $k$ dimensional row vector of $0$ resp $1$. The output will be computed on the token stream of $\bm{e}_2$.

\paragraph{Linear attention} First, the attention layer will compute the forward pass $\bm{W}(\bm{z}) \bm{x}$. To do this, let us fix $H=M$ heads, $d_q=d_k=1$ and $d_v=h$. For each head $m$, we can construct the value matrix such that the first token has a value vector $\bm{\theta}_m\bm{x}$ while the second has $\bm{0}_h$. By choosing the key and query matrices correctly, the attention score between the first and second token can be made to be exactly $z_m$. By letting the projection matrix be constant across heads, the attention operation would then be 
\begin{equation}
\bm{e}_2 \leftarrow \bm{e}_2 + \sum_{m}^M z_m(\bm{\theta}_m\bm{x})^\top \bm{W}_P    
\end{equation}
by appropriately choosing $\bm{W}_P$ the residual stream would then equal $(\bm{0}_d,\bm{z}^\top,\bm{W}(\bm{z})\bm{x},\bm{0}_{o})^\top$ after the attention layer.

\paragraph{MLP} Finally, the MLP layer simply applies the correct nonlinearity $\sigma$ to $\bm{W}(\bm{z})\bm{x}$ and applies the readout weight $\bm{A}$ to finally write the result on the remaining $\bm{0}_o$ in the residual stream.

\section{Additional results}

\begin{figure}[h]
    \centering
    \includegraphics{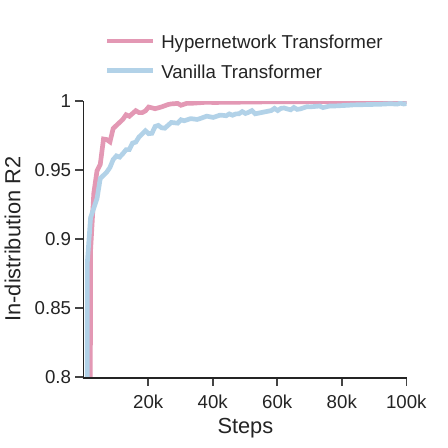}
    \caption{Performance on in-distribution tasks.}
    \label{appfig:test-r2}
\end{figure}

\section{Experimental details}

\subsection{Data generation} \label{app:task_generation}

The data is generated using a teacher hypernetwork. We first initialize the teacher parameters once. Then, for each sequence, we sample a task latent variable $\bm{z}$ which induces a noiseless mapping from inputs $\bm{x}$ to a scalar target $y$ following equation 
\begin{align}
y = \bm{a}^\top \sigma(\bm{W}(\bm{z}) \bm{x}),
\end{align}
where $\sigma$ is the GELU nonlinearity.

The weight $\bm{W}$ is the linear combination of modules $\{\bm{\theta}_m\}_m$ by $\bm{z}$, i.e. $\bm{W}(\bm{z}) = \sum_m \bm{\theta}_m z_m$. In order to make sure the forward pass is well-behaved, we furthermore normalize the generated weight $\bm{W}$ by its operator norm.

For all experiments, we fix the task latent variable dimension to $M=6$, input dimension to $d = 16$, hidden dimension of the teacher to $h=16$, and output dimension to $o=1$.
The teacher parameters $\{\bm{\theta}_m\}_m$ and $\bm{a}$ are generated by sampling the entries i.i.d. from the centered truncated normal distribution, with standard deviation resp. $\frac{1}{\sqrt{M}}$ and $\frac{1}{\sqrt{h}}$.
We define the distribution over inputs $\bm{x}$ to be the uniform distribution with $0$ mean and standard deviation $1$ on $\mathbb{R}^d$. 

Finally, we specify the distribution over task latent variable $\bm{z}$.

\paragraph{Task latent variable distribution.} 
Here, we consider tasks where modules are sparsely, and linearly combined.
A task distribution is specified by a set of masks, that are binary vectors of $\mathbb{R}^M$. 
Given a mask, we sample a task $\bm{z}$ as follows. We first sample an $M$-dimensional random variable following the exponential distribution. Then, we zero out entries corresponding to the mask being $0$. We normalize the vector such that the sum equals 1. This procedure simulates the uniform sampling from the simplex spanning the directions in which the mask is non-zero. Finally, we add the mask to the vector and rescale the outcome by $0.5$. This ensures two tasks generated by distinct masks do not have intersecting support (but intersecting span). See Algorithm \ref{alg:task_sample} for the pseudocode.

\begin{algorithm}
\caption{Algorithm to sample the test latent variable from given a mask.}
\label{alg:task_sample}
\begin{algorithmic}
\Require mask $\bm{m}$ of size $M$
\Return task latent variable $\bm{z}$
\State sample $M$-dimensional vector $\bm{z}$ from the exponential distribution
\State $\bm{z} \gets \bm{z} \odot \bm{m}$
\State $\bm{z} \gets \frac{\bm{z}}{\|\bm{z}\|_1}$
\State $\bm{z} \gets 0.5 \cdot (\bm{z} + \bm{m})$
\State return $\bm{z}$
\end{algorithmic}
\end{algorithm}

The task distribution is then generated as follows: first, a mask is sampled randomly and uniformly from the prespecified set. Then, the vector $\bm{z}$ is sampled following the above procedure.

\paragraph{Connected and disconnected task support}
\label{app:connectedness}

Controlling the task distribution in this way allows us to study under what circumstances it is possible to generalize to the full support after having only observed demonstrations from a subset of tasks.
More precisely, if $\mathcal{P}_z$ is a distribution on the latent code that does not have full support on $\mathbb{R}^M$, can a system trained only on tasks sampled from $\mathcal{P}_z$  generalize to the full space?
Here, we assume that the support of $\mathcal{P}_z$ spans the whole space $\mathbb{R}^M$.
We will investigate two situations: when $\mathcal{P}_z$ has \textit{connected} support and when it has \textit{disconnected} support.
For a formal definition, we defer the reader to \cite{schug_discovering_2024}.
Intuitively, having connected support means that no subset of modules appears solely in isolation of the rest.
To make a concrete example for the simple case of $M=3$ modules, if the support of $\mathcal{P}_z$ is $\left(\mathbb{R}^2\times \{0\}\right) \cup \left(\{0\}^2 \times \mathbb{R}\right)$, the learner will have never seen the interaction of the first two modules with the last one and hence the support is disconnected.
\citep{schug_discovering_2024} theoretically show that when the support is disconnected, there are several failure cases impeding compositional generalization.

\subsection{Training and Evaluation metrics} \label{app:metric}

In all our experiments, we train the model on tasks generated from a set of binary masks as described in Section ~\ref{app:task_generation}. During training, both $\bm{x}$ and $\bm{z}$ are sampled online. For experiments investigating the effect of connected and disconnected task support (Panel D of Figure~\ref{fig:results}), we train the same model on the masks listed in the corresponding columns in Table~\ref{tab:task_distribution}, where we make sure the same number of tasks is used during training in both settings. For all other experiments, we use the masks of \textit{Connected+} during training.

\paragraph{OOD $R^2$}
To evaluate the performance of the models on compositional generalization, we compute the $R^2$ score of the linear regression on tasks generated by 2-hot masks that were unseen during training. For example, for models trained on tasks from masks in \textit{Connected+} (c.f. Table~\ref{tab:task_distribution}), the OOD evaluation is done on tasks from masks $(1,0,0,1,0,0),(0,1,0,0,1,0),(0,0,1,0,0,1)$.

Given a sequence of $N-1$ pairs $(\bm{x}_i, y_i)$ and a prediction $\tilde{y}$ for $y_{N}$, the $R^2$ score is defined as the MSE loss between $y_N$ and $\tilde{y}$, normalized by the MSE loss between $y_{N}$ and $\frac{1}{N-1}\sum_i^{N-1} y_i$. The score is averaged over 16000 OOD sequences.

\paragraph{Probing latent}

In order to probe whether the transformer implicitly learned to infer the latent task variable $\bm{z}$, we linearly probe the residual stream throughout training. Given a model, we collect a batch of 16000 sequences $\bm{\mathrm{X}}$ sampled from the training task distribution, associated to various task latent variables $\bm{\mathrm{Z}}$. We then train a linear regressor from $\bm{\mathrm{X}}$ to $\bm{\mathrm{Z}}$ by Ridge regression, with regularization strength 1. Then, we evaluate the $R^2$ score of the regressor on 16000 sequences drawn from the OOD distribution, against their respective latent code.

\paragraph{Random unstructured control}
\label{app:random-control-task}
A lazy way for a model to learn the task is to ignore its compositional structure, and simply infer the target solely based on the context of the current task. If that is the case, then the model should be able to have a reasonable guess of $g(\bm{x}_N,\bm{\omega})$ when the context provides demonstrations $(\bm{x}_i, g(\bm{x}_i, \bm{\omega}))$, with $\bm{\omega} \not\in \{\bm{W}(\bm{z}) \mid \bm{z} \in\mathbb{R}^M\}$.
We evaluate our models on such an unstructured control task where $\bm{\omega}$ is sampled as the output of a freshly initialized hypernetwork teacher and random latent code, both of which share no other structure with the training tasks.

\begin{table}[]
\centering
\addtolength{\tabcolsep}{-3pt}
\caption{Training task distribution used for our experiments. Each column defines a distribution. The vectors specify the set of masks used in the training set to generate the distribution. Both \textit{Connected} and \textit{Connected+} task support have connected support, while \textit{Disconnected} have disconnected support.}
\label{tab:task_distribution}
\begin{tabular}{lll}
\toprule 
Connected & Disconnected & Connected+ \\ \midrule
(1,1,0,0,0,0) & (1,1,0,0,0,0) & (1,1,0,0,0,0) \\
(0,1,1,0,0,0) & (0,1,1,0,0,0) & (0,1,1,0,0,0) \\
(0,0,1,1,0,0) & (1,0,1,0,0,0) & (0,0,1,1,0,0) \\
(0,0,0,1,1,0) & (0,0,0,1,1,0) & (0,0,0,1,1,0) \\
(0,0,0,0,1,1) & (0,0,0,0,1,1) & (0,0,0,0,1,1) \\
(1,0,0,0,0,1) & (0,0,0,1,0,1) & (1,0,0,0,0,1) \\
  &   & (1,0,1,0,0,0) \\
  &   & (0,1,0,1,0,0) \\
  &   & (0,0,1,0,1,0) \\
  &   & (0,0,0,1,0,1) \\
  &   & (1,0,0,0,1,0) \\
  &   & (0,1,0,0,0,1) \\
\bottomrule          
\end{tabular}

\end{table}

\subsection{Architecture}
\label{app:architecture}
\subsubsection{Vanilla transformer}
The input is $$\bm{\mathrm{X}} = ((\bm{x}_1, y_1), \ldots, (\bm{x}_{N-1}, y_{N-1}), (\bm{x}_N, 0))$$
where $\bm{x}_N$ is the "query" input whose image we have to infer.

The vanilla transformer consists of a standard decoder-only multi-layer transformer, where each block is structured as
\begin{align*}
    \bm{\mathrm{X}} &\leftarrow \texttt{MultiHeadAttention}(\texttt{LayerNorm}(\bm{\mathrm{X}})) + \bm{\mathrm{X}} \\
    \bm{\mathrm{X}} &\leftarrow \texttt{Feedforward}(\texttt{LayerNorm}(\bm{\mathrm{X}})) + \bm{\mathrm{X}}.
\end{align*}
\texttt{MultiHeadAttention} is multi-head softmax attention and uses T5-style relative positional embeddings \citep{raffel2020exploring}. 
The feedforward layer is a two-layer MLP with GELU nonlinearity. It is applied to each token in the sequence independently.

A final readout layer projects the query token to the output dimension.

\subsubsection{Transformer Hypernetwork}
The Transformer Hypernetwork is a single hidden layer MLP whose weights are generated by a transformer. More precisely:
\begin{itemize}
    \item the first layer weights are generated by a vanilla transformer. It gets as input the sequence 
    $$\bm{\mathrm{X}} = ((\bm{x}_1, y_1), \ldots, (\bm{x}_N, y_N), (\bm{0}, 0))$$
    The output of the last token is projected to a latent code space $\mathbb{R}^{\hat{m}}$, followed by a readout to the dimension of the weight matrix of the first MLP layer. One can reinterpret this as the transformer generating a latent code $\bm{z}$, and then generating the weight matrix $\bm{W}(\bm{z}) = \sum_{m=1}^{\hat{M}} z_m \bm{\theta}_m$, where the $\bm{\theta}_m$ are the learned modules.
    \item the readout weights are learnable parameters (i.e. not generated by the transformer)
\end{itemize}

\subsection{Hyperparameters}

\begin{table}[H]
\begin{center}
\caption{Implementation and optimization details.}
\vspace{0.5cm}
\begin{tabular}{@{}ll@{}}
\toprule
Positional embedding & Relative \\
Attention mask & None \\
Optimizer &  AdamW \\
Scheduler &  Cosine annealing\\
Number of gradient steps & 100 000  \\
\bottomrule
\end{tabular}
\end{center}
\end{table}

We selected the following hyperparameter based on the mean OOD $R^2$ score on 3 seeds:
\begin{table}[H]
\begin{center}
\caption{Hyperparameters.}
\footnotesize
\vspace{0.5cm}
\begin{tabular}{@{}lccl@{}}
\toprule
Hyperparameter & Vanilla Transformer & Hypernetwork Transformer & Sweep values \\
\toprule
Embedding dim &128&64& 32, 64, 128, 256\\
Number of heads &4&4& 1, 2, 4\\
Number of layers &2&2& 1, 2, 3, 4, 6\\
Expansion factor of FFN &4&4& 4\\
Hypernetwork latent dim & NA &6& 6,8,12\\
Main MLP hidden layer dim & NA & 32& 16, 32\\\hline
Learning rate &0.001&0.001& 0.003, 0.001, 0.0003\\
Weight decay & 0.1&0& 0, 0.1\\
Gradient clipping &1&1&  1\\
Batch size &128&128& 128\\
\bottomrule
\end{tabular}
\end{center}
\end{table}

\end{document}